\documentclass[letterpaper, 10 pt, conference]{ieeeconf}
\IEEEoverridecommandlockouts                              
\usepackage{times}
\usepackage{epsfig}
\usepackage{graphicx}
\usepackage{multirow}
\usepackage{multicol}

\usepackage{comment}
\usepackage{url}

\usepackage{graphicx}
\usepackage{epsfig}
\usepackage{epic,eepic}
\usepackage{times}
\usepackage{mathptmx}
\usepackage{cite}
\usepackage{color}

\usepackage{times}

\definecolor{red}{rgb}{1,0,0}
\definecolor{green}{rgb}{0,1,0}
\definecolor{blue}{rgb}{0,0,1}
\definecolor{violet}{rgb}{1,0,1}
\definecolor{cyan}{cmyk}{1,0,0,0}
\definecolor{magenta}{cmyk}{0,1,0,0}
\definecolor{yellow}{cmyk}{0,0,1,0}

\definecolor{white}{rgb}{1,1,1}

\usepackage{arydshln}

\newcommand{\CO}[1]{}

\newcommand{\CommentOut}[1]{}

 \newcommand{\editage}[1]{}

\begin{document}

\title{\LARGE \bf
Active Semantic Localization
with
Graph Neural Embedding
}

\author{%
Mitsuki Yoshida$^{*}$, Kanji Tanaka$^{*}$, Ryogo Yamamoto$^{*}$, and Daiki Iwata$^{*}$
\thanks{$*$M. Yoshida, K. Tanaka, R. Yamamoto, and D. Iwata are with Department of Engineering, University of Fukui, Japan. {\tt\small tnkknj@u-fukui.ac.jp}}}

\maketitle


\newcommand{\TAB}[1]{#1}

\newcommand{\FIG}[3]{
\begin{minipage}[b]{#1cm}
\begin{center}
\includegraphics[width=#1cm]{#2}\\
{\scriptsize #3}
\end{center}
\end{minipage}
}

\newcommand{\FIGU}[3]{
\begin{minipage}[b]{#1cm}
\begin{center}
\includegraphics[width=#1cm,angle=180]{#2}\\
{\scriptsize #3}
\end{center}
\end{minipage}
}

\newcommand{\FIGm}[3]{
\begin{minipage}[b]{#1cm}
\begin{center}
\includegraphics[width=#1cm]{#2}\\
{\scriptsize #3}
\end{center}
\end{minipage}
}

\newcommand{\FIGR}[3]{
\begin{minipage}[b]{#1cm}
\begin{center}
\includegraphics[angle=-90,width=#1cm]{#2}
\\
{\scriptsize #3}
\vspace*{1mm}
\end{center}
\end{minipage}
}

\newcommand{\FIGRpng}[5]{
\begin{minipage}[b]{#1cm}
\begin{center}
\includegraphics[bb=0 0 #4 #5, angle=-90,clip,width=#1cm]{#2}\vspace*{1mm}
\\
{\scriptsize #3}
\vspace*{1mm}
\end{center}
\end{minipage}
}

\newcommand{\FIGCpng}[5]{
\begin{minipage}[b]{#1cm}
\begin{center}
\includegraphics[bb=0 0 #4 #5, angle=90,clip,width=#1cm]{#2}\vspace*{1mm}
\\
{\scriptsize #3}
\vspace*{1mm}
\end{center}
\end{minipage}
}

\newcommand{\FIGpng}[5]{
\begin{minipage}[b]{#1cm}
\begin{center}
\includegraphics[bb=0 0 #4 #5, clip, width=#1cm]{#2}\vspace*{-1mm}\\
{\scriptsize #3}
\vspace*{1mm}
\end{center}
\end{minipage}
}

\newcommand{\FIGtpng}[5]{
\begin{minipage}[t]{#1cm}
\begin{center}
\includegraphics[bb=0 0 #4 #5, clip,width=#1cm]{#2}\vspace*{1mm}
\\
{\scriptsize #3}
\vspace*{1mm}
\end{center}
\end{minipage}
}

\newcommand{\FIGRt}[3]{
\begin{minipage}[t]{#1cm}
\begin{center}
\includegraphics[angle=-90,clip,width=#1cm]{#2}\vspace*{1mm}
\\
{\scriptsize #3}
\vspace*{1mm}
\end{center}
\end{minipage}
}

\newcommand{\FIGRm}[3]{
\begin{minipage}[b]{#1cm}
\begin{center}
\includegraphics[angle=-90,clip,width=#1cm]{#2}\vspace*{0mm}
\\
{\scriptsize #3}
\vspace*{1mm}
\end{center}
\end{minipage}
}

\newcommand{\FIGC}[5]{
\begin{minipage}[b]{#1cm}
\begin{center}
\includegraphics[width=#2cm,height=#3cm]{#4}~$\Longrightarrow$\vspace*{0mm}
\\
{\scriptsize #5}
\vspace*{8mm}
\end{center}
\end{minipage}
}

\newcommand{\FIGf}[3]{
\begin{minipage}[b]{#1cm}
\begin{center}
\fbox{\includegraphics[width=#1cm]{#2}}\vspace*{0.5mm}\\
{\scriptsize #3}
\end{center}
\end{minipage}
}









\newcommand{\tabA}{
\begin{table*}
\begin{center}
\caption{Performance results.}\label{tab:A}
\TAB{
\begin{tabular}{|l|l|rrrr|rrrr|rrrr|}
\hline
\multicolumn{2}{|l|}{Domain-gap} &
\multicolumn{4}{l}{Dataset x Method} &
\multicolumn{8}{r|}{} \\
\hline
\multirow{2}{3em}{$T_{xy}$} & \multirow{2}{3em}{$T_{\theta}$} &
\multicolumn{4}{r|}{00800-TEEsavR23oF} &
\multicolumn{4}{r|}{00801-HaxA7YrQdEC} &
\multicolumn{4}{r|}{00809-Qpor2mEya8F} \\
& & ~~SV & ~~MV & ~Ours & ~~DA & ~~SV & ~~MV & ~Ours & ~~DA & ~~SV & ~~MV & ~Ours & ~~DA \\
\hline
\multirow{5}{3em}{1.0} & 90 & 38 & 52 & 61 & 54 & 47 & 49 & 52 & 53 & 43 & 41 & 52 & 49 \\
 & 70 & 38 & 53 & 62 & 52 & 36 & 52 & 55 & 50 & 43 & 48 & 52 & 52 \\
 & 50 & 45 & 52 & 58 & 57 & 43 & 51 & 57 & 60 & 37 & 54 & 54 & 54 \\
 & 30 & 46 & 54 & 57 & 58 & 48 & 54 & 60 & 60 & 43 & 51 & 53 & 57 \\
 & 10 & 43 & 55 & 54 & 51 & 45 & 54 & 58 & 67 & 42 & 52 & 59 & 58 \\
\hline
\multirow{5}{3em}{0.8} & 90 & 52 & 62 & 66 & 70 & 46 & 52 & 56 & 52 & 44 & 54 & 55 & 56 \\
 & 70 & 46 & 57 & 55 & 57 & 39 & 53 & 49 & 57 & 50 & 48 & 53 & 53 \\
 & 50 & 38 & 50 & 56 & 56 & 49 & 53 & 54 & 53 & 40 & 49 & 54 & 49 \\
 & 30 & 40 & 54 & 54 & 53 & 43 & 50 & 58 & 62 & 33 & 44 & 47 & 44 \\
 & 10 & 47 & 59 & 66 & 68 & 42 & 57 & 53 & 53 & 36 & 49 & 46 & 48 \\
\hline
\multirow{5}{3em}{0.6} & 90 & 39 & 46 & 43 & 42 & 41 & 56 & 58 & 62 & 37 & 45 & 45 & 47 \\
 & 70 & 42 & 56 & 53 & 49 & 49 & 53 & 62 & 61 & 37 & 40 & 44 & 48 \\
 & 50 & 42 & 51 & 56 & 57 & 46 & 51 & 51 & 56 & 46 & 59 & 51 & 50 \\
 & 30 & 48 & 53 & 55 & 59 & 46 & 49 & 52 & 53 & 35 & 46 & 52 & 52 \\
 & 10 & 33 & 47 & 49 & 41 & 34 & 51 & 55 & 61 & 50 & 56 & 55 & 53 \\
\hline
\multirow{5}{3em}{0.4} & 90 & 33 & 49 & 51 & 52 & 43 & 49 & 55 & 54 & 47 & 48 & 53 & 56 \\
 & 70 & 42 & 55 & 56 & 52 & 43 & 47 & 60 & 59 & 44 & 48 & 49 & 47 \\
 & 50 & 48 & 57 & 60 & 57 & 42 & 55 & 59 & 60 & 40 & 50 & 48 & 53 \\
 & 30 & 46 & 57 & 54 & 53 & 40 & 46 & 48 & 51 & 38 & 50 & 60 & 58 \\
 & 10 & 40 & 49 & 54 & 50 & 47 & 53 & 54 & 52 & 34 & 54 & 57 & 53 \\
\hline
\multirow{5}{3em}{0.2} & 90 & 36 & 54 & 56 & 57 & 46 & 60 & 60 & 60 & 41 & 48 & 50 & 51 \\
 & 70 & 50 & 60 & 63 & 60 & 41 & 51 & 52 & 52 & 48 & 53 & 49 & 47 \\
 & 50 & 51 & 59 & 60 & 62 & 50 & 54 & 54 & 56 & 34 & 46 & 51 & 48 \\
 & 30 & 38 & 47 & 57 & 55 & 46 & 59 & 55 & 55 & 40 & 50 & 45 & 48 \\
 & 10 & 49 & 55 & 57 & 53 & 43 & 53 & 56 & 51 & 33 & 53 & 56 & 54 \\
\hline
\end{tabular}
}
\end{center}
\end{table*}
}

\newcommand{\figETM}{
\begin{figure}[t]
\FIG{8}{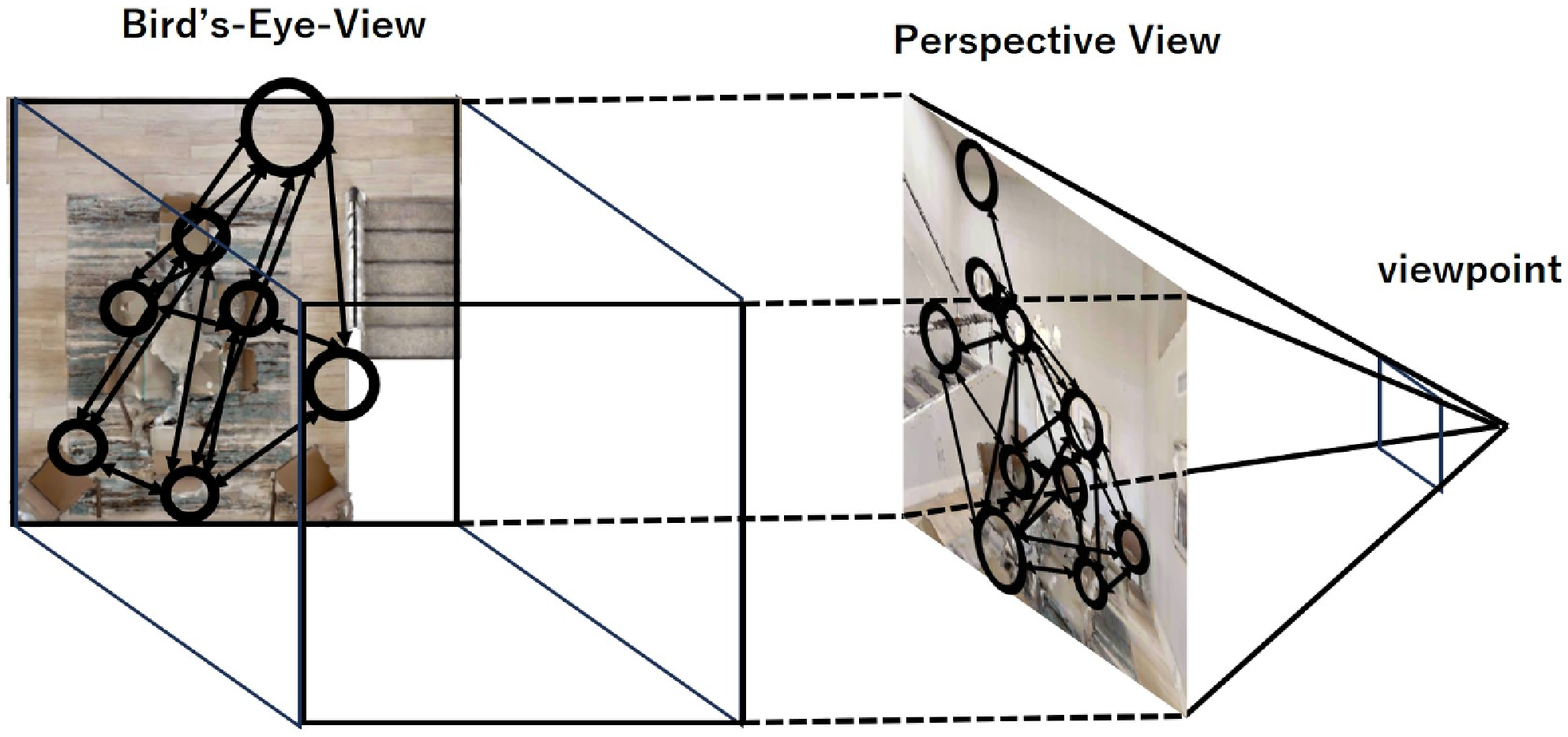}{}
\caption{%
Topological navigation using ego-centric topological maps.
Left: Conventional world-centric map.
Right: The proposed ego-centric map.
}\label{fig:ETM}
\end{figure}
}

\newcommand{\figA}{
\begin{figure}
\begin{center}
\vspace*{-5mm}
\hspace*{2mm}
\FIG{8.5}{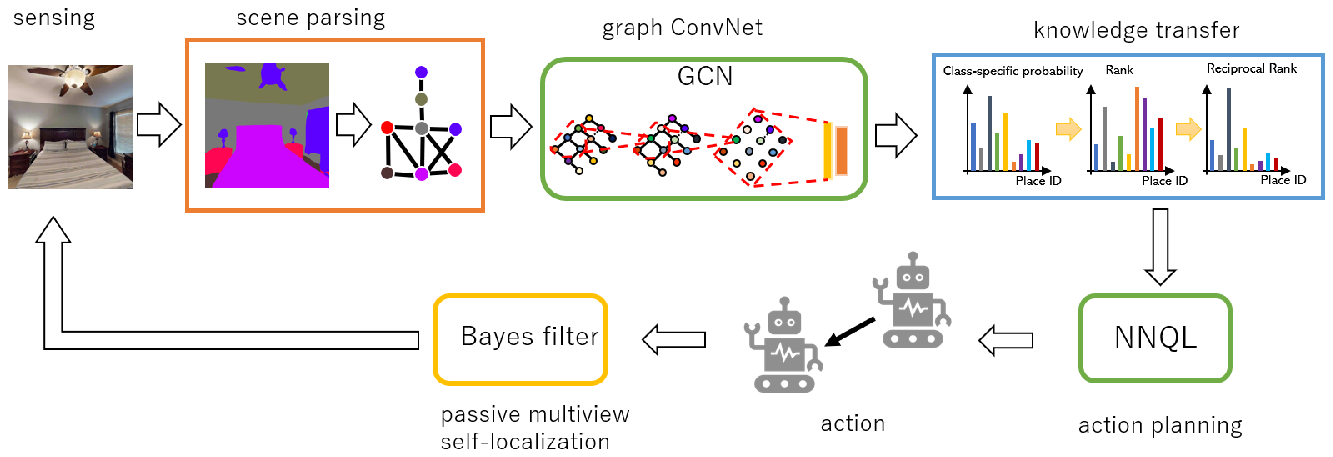}{}
\end{center}
\caption{Our framework generates a scenegraph, embeds it, and transfers it to the active planner.}\label{fig:A}
\vspace*{-5mm}
\end{figure}
}

\newcommand{\figB}{
\begin{figure}
\begin{center}
\vspace*{-5mm}
\FIG{8}{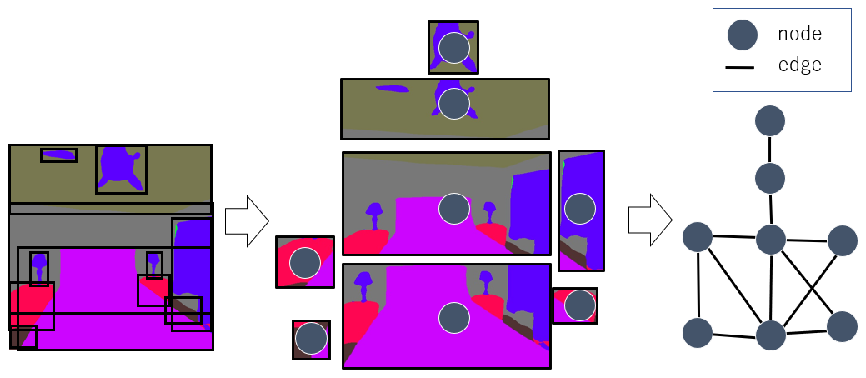}{}
\end{center}
\caption{Semantic scene graph.}\label{fig:B}
\end{figure}
}

\newcommand{\figD}{
\begin{figure}
\begin{center}
\vspace*{-5mm}
\hspace*{4mm}
\FIG{8.2}{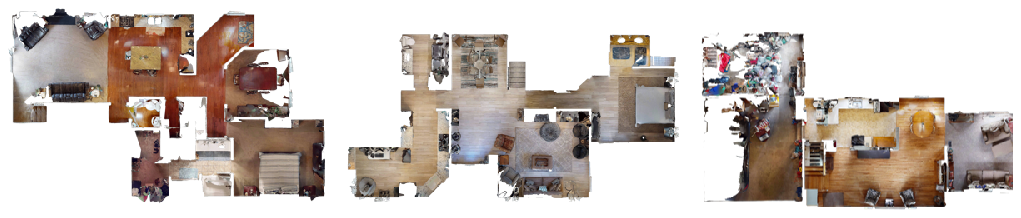}{}\\
\FIG{8.2}{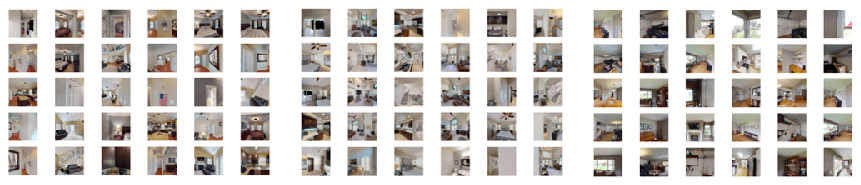}{}
\end{center}
\caption{The robot workspaces and examples images for ``00800-TEEsavR23oF," ``00801-HaxA7YrQdEC," and ``00809-Qpor2mEya8F" datasets.}\label{fig:D}
\end{figure}
}

\newcommand{\figE}{
\begin{figure}
\begin{center}
\vspace*{-5mm}
\hspace*{2mm}
\FIG{8.5}{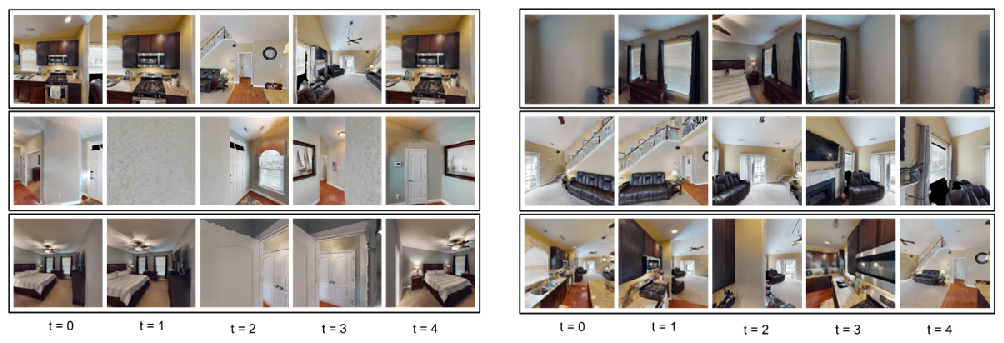}{}\\
\scriptsize
\end{center}
\caption{Examples of view sequences.
Left and right panels show successful and unsuccessful cases of self-localization, respectively.}\label{fig:E}
\end{figure}
}

\newcommand{\figF}{
\begin{figure}
\begin{center}
\hspace*{2mm}
\FIGR{8.5}{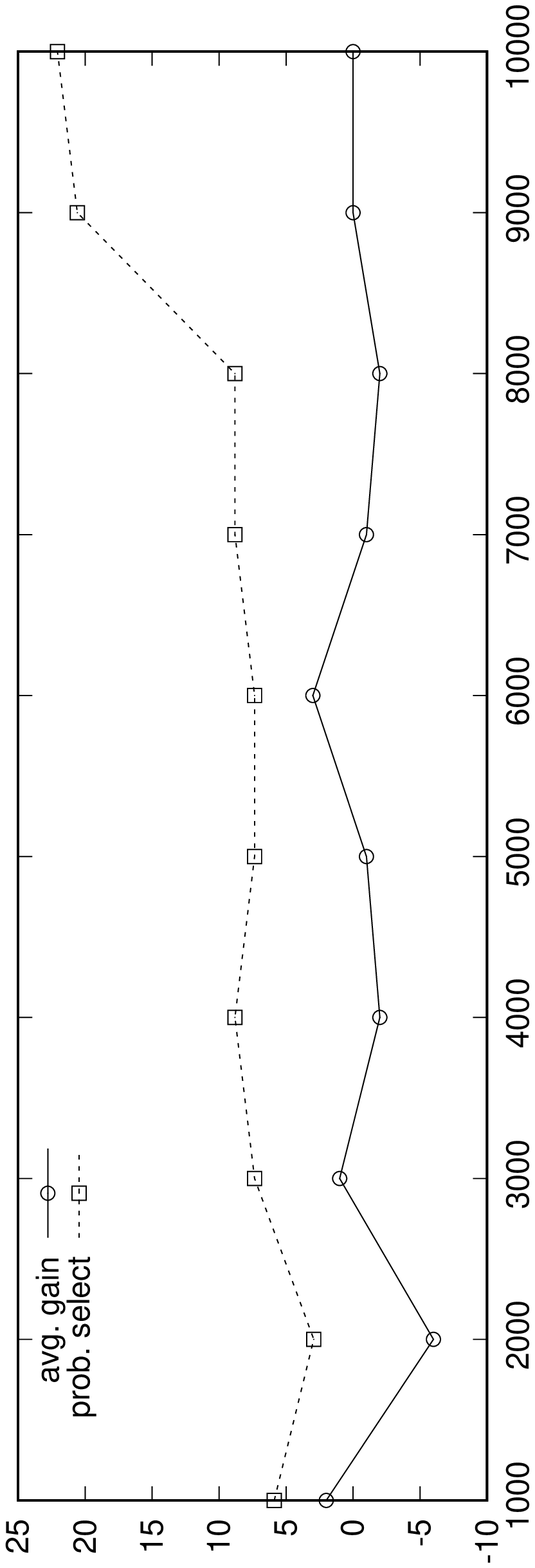}{}
\end{center}
\vspace*{-5mm}
\caption{Results in unsupervised domain adaptation (UDA) applications.}\label{fig:F}
\end{figure}
}

\begin{abstract}
Semantic localization, i.e., robot self-localization with semantic image modality, is critical in recently emerging embodied AI applications (e.g., point-goal navigation, object-goal navigation, vision language navigation) and topological mapping applications (e.g., graph neural SLAM, ego-centric topological map). However, most existing works on semantic localization focus on passive vision tasks 
without viewpoint planning, or rely on additional rich modalities (e.g., depth measurements). Thus, the problem is largely unsolved. In this work, we explore a lightweight, entirely CPU-based, domain-adaptive semantic localization framework, called graph neural localizer. 
Our approach is inspired by two recently emerging technologies: (1) Scene graph, which combines the viewpoint- and appearance- invariance of local and global features; (2) Graph neural network, which enables direct learning/recognition of graph data (i.e., non-vector data). 
Specifically, a graph convolutional neural network is first trained as a scene graph classifier for passive vision, and then its knowledge is transferred to a reinforcement-learning planner for active vision. 
Experiments on two scenarios, self-supervised learning and unsupervised domain adaptation, using a photo-realistic Habitat simulator validate the effectiveness of the proposed method.
\end{abstract}

\begin{keywords}
graph neural embeddings,
active semantic localization,
knowledge transfer,
domain adaptation
\end{keywords}

\section{Introduction}

Semantic localization, i.e., robot self-localization with semantic image modality, is critical in recently emerging embodied AI applications \cite{habitatchallenge} (e.g., point-goal navigation \cite{pgn}, object-goal navigation \cite{ogn}, vision language navigation \cite{vln}) and topological mapping applications (e.g., graph neural SLAM, ego-centric topological map). It has become clear that the performance of semantic localization is the bottleneck in these navigation applications \cite{pgnloc}. 

Most existing works on semantic localization focus on passive vision tasks without viewpoint planning \cite{vsl}, 
or rely on additional rich modalities (e.g., depth measurements) \cite{sem3d}. Thus, the problem is largely unsolved. Furthermore, current methods rely on expensive GPU environments to train the deep learning models for state recognition \cite{nbnnlearning} 
and action planning \cite{anl}. 
This limits their application domains and for example, it makes them unapplicable to lightweight robotics applications such as household personal robots \cite{100dollar}. 

Motivated by these challenges, in this work we present a lightweight, entirely CPU-based, domain-adaptive semantic localization framework, called graph neural localizer. 
Our approach is inspired by two recently emerging technologies: (1) Scene graph \cite{sgsurvey}, which combines the viewpoint- and appearance- invariance of local and global features; 
(2) Graph neural network \cite{gnnsurvey}, which enables direct learning/recognition of graph data (i.e., non-vector data). Specifically, a graph convolutional neural network is first trained as a scene graph classifier for passive vision, and then its knowledge is transferred to a next-best-view planner for active vision. Although such a knowledge transfer task from passive to active self-localization has been recently addressed 
for conventional models \cite{cvmi2022kurauchi}, 
it has not been sufficiently explored for the 
recently-emerging graph neural network models. To this end, detailed implementation issues including scene parsing \cite{sggsurvey}, 
unsupervised knowledge transfer \cite{rrf2009}, 
and efficient reinforcement-learning \cite{nnql}
are 
implemented, evaluated and 
discussed. Experiments on two scenarios, self-supervised learning \cite{sslsurvey} 
and unsupervised domain adaptation \cite{fsda}, 
using a photo-realistic Habitat simulator \cite{habitatsim} 
validate the effectiveness of the proposed method.

The contributions of this research are summarized as follows:
(1)
An entirely CPU-based lightweight framework for solving semantic localization in both active and passive 
vision tasks is presented.
(2)
A scene graph and a graph neural network are combined for the first time to solve semantic localization, presenting a new 
graph neural localizer framework.
(3)
Experiments show that the proposed method outperforms the baseline method in terms of self-localization performance, computational efficiency, and domain adaptation.

\figA

\section{
Related Work
}

Visual robot self-localization has been extensively studied in various formulations \cite{deepvprsurvey} 
such as image retrieval \cite{fabmap}, 
multi-hypothesis tracking \cite{mcl}, 
geometric matching \cite{vsl}, 
place classification \cite{planet}, 
and viewpoint regression \cite{regress}. 
It is also closely related to the task of 
loop closure detection \cite{lcdsurvey}, 
an essential component of visual robot mapping.
This work focuses on the classification formulation, where supervised and self-supervised learning of image classifier are directly applicable and have become a predominant approach \cite{icra2019sc}.

Existing self-localization approaches are broadly divided into two approaches, local and global feature \cite{vprsurvey}, 
depending on the type of scene model. The local feature approach describes an image in a viewpoint-invariant manner using a ``bag" of keypoint-level features \cite{ibowlcd}. 
The global feature approach describes an image in an appearance-invariant manner using a single image-level feature \cite{global}. 
There is also a hybrid approach called part feature \cite{partfeat}, 
in which scenes are described in a viewpoint-invariant and appearance-invariant manner using subimage-level features. The scene graph model \cite{sgsurvey} 
used in this research can be viewed as an extension of the part feature, 
in that 
it can describe not only part features by graph nodes,
but also the relationships between parts by graph edges.

Appearance features, manually designed or 
deeply learned, are a predominant approach for in-domain self-localization \cite{lcdsurvey}. %
In contrast, the semantic feature used in this study 
has proven to be advantageous in cross-domain setups (e.g., cross-modality \cite{crossmodalsem}, 
cross-view \cite{crossviewsem}, 
cross-dataset \cite{crossdssem}) 
as in \cite{vsl}, which was enabled by recent advances in deep semantic imaging technology \cite{crossdssem}.

Most existing works on semantic localization focus on passive self-localization tasks without 
viewpoint planning \cite{vsl}, or rely on additional rich modalities (e.g., 3d point clouds) 
\cite{sem3d}. 
In contrast, 
in our framework,
both the passive and active self-localization tasks
are addressed,
and
both training and deployment stages do not rely on other modalities.
Note that 
semantic localization can be ill-posed for a passive observer, as many viewpoints may not provide any discriminative semantic feature. The active self-localization aims to adapt the observer's viewpoint trajectory, avoiding non-salient scenes that do not provide a landmark view or moving efficiently toward places that are likely to be most informative with the aim of reducing sensing and computation costs. Most existing works on active localization focus on rich modalities such as RGB image \cite{dal}
and 3D point clouds \cite{anl}.
In contrast,
the issue of
active semantic localization 
has not been sufficiently explored.

Active self-localization is also related to other active vision tasks, including first-person-view semantic navigation (e.g., point-goal navigation \cite{pgn}, vision-language navigation \cite{vln}, 
object-goal navigation \cite{ogn}), which has recently emerged in the robotics and vision communities. However, 
most existing works assume the availability 
of ground-truth viewpoint 
information during training episodes.
In contrast,
the availability of such rich ground-truth is not assumed 
in this work,
but 
only a sparse reward 
given upon successful localization 
is assumed as supervision.

As described above, the issues of 
visual place classification, scene graph, graph neural networks, and active vision have been researched independently. This work is the first to bring together these approaches to address the emerging issue of semantic localization.

\figB

\section{Problem}\label{sec:prob}

The problem of active self-localization
is formulated as 
an instance of discrete-time discounted Markov decision process (MDP)
where agents 
(i.e., robots)
interact with probabilistic environments \cite{suttonrl}. 
A time-discounted MDP 
is a general formulation,
consisting of
the state set $S$,
the action set $A$,
the state transition distribution $P$,
the reward function $R$,
and the discount rate $\gamma$.
In our specific scenario,
$S$ is 
a set of 
embeddings of 
first-person-view images,
$A$ is a set of possible 
rotating $r$
or
forward
$f$
motions,
and 
$R$ is the reward given upon successful localization.

The application domain of active 
self-localization covers various sensor modalities (e.g., monocular cameras, LRF, sonar), workspaces (e.g., university campus, office, underwater, airborne) and various types of domains (e.g., object placement, viewpoint trajectories, weathers, lighting conditions). Nevertheless, rich training environments are provided only for very limited application scenarios. The Habitat simulator \cite{habitatsim}, which provides photo-realistic RGB images, is one of such valuable training environments in embodied AI. 
Therefore, it is employed in this work.

Semantic images are generated from the RGB images by using an independent off-the-shelf model for semantic segmentation, as detailed in Section \ref{sec:ssg}. Although 
ideal low-noise semantic images
could be provided by the Habitat simulator,
we do not rely on them 
to focus on the more realistic setting.

Performance of an active vision 
system 
should be measured in terms of its generalization ability \cite{genabi}. 
To do so, it is important to have a domain gap 
between the training and test domains. In this work, 
we consider a specific type of cross-domain scenario called ``cross-view localization" 
where domain gap is defined as the difference in the distribution of start viewpoints between 
the training and deployment domains. 
Specifically, 
the amount of domain gap
is 
controlled by two preset parameters,
called 
location difference $T_{xy}$ and bearing difference $T_{\theta}$.
Given 
$T_{xy}$ and $T_{\theta}$,
the training and test sets of start viewpoints are sampled in the following procedure. 
First, the test set $\{(x_{test}^i$, $y_{test}^i$, $\theta_{test}^i)\}$ is uniformly sampled from the robot workspace. Then, the training samples $\{(x, y, \theta)\}$ 
is sampled by iteratively
sampling a location 
$(x, y)$
that satisfies the condition $((x_{test}^i-x)^2$$+(y_{test}^i-y)^2)^{1/2}$$>$$T_{xy}$, 
and then checking
the bearing condition  $|\angle(\theta_{test}^i$$-$$\theta)|$$>$$T_{\theta}$ between 
it
and its nearest-neighbor test sample.

In experiments,
25 different settings of 
location and bearing difference
$(T_{xy}, T_{\theta})$,
which are 
combinations
of 5 different settings of location difference
$T_{xy}$
($\in\{$0.2, 0.4, 0.6, 0.8, 1.0$\}$),
and
5 different settings of bearing difference
$T_{\theta}$ 
($\in\{$10, 30, 50, 70, 90$\}$)
will be considered.

\section{
Approach
}

Figure \ref{fig:A} shows the semantic localization framework. It consists of three main modules: (1) a scene parser module that parses an image into a semantic scene graph, (2) a graph embedding module that embeds a semantic scene graph into a state vector, and (3) an action planner module that maps the state vector to an action plan. Each module is detailed in the subsequent subsections.

\subsection{
Semantic Scene Graph
} \label{sec:ssg}

Similarity-preserving image-to-graph mapping is one of most essential requirements for scene graph embedding used by 
an active vision. It is 
particularly
necessary for 
the agent's similar behavior to be reproduced in different domains. Most existing models such as deep learning -based scene graph generation \cite{sggsurvey} 
are not aimed at 
localization applications 
and 
thus do not have the 
similarity preserving ability. 

As an alternative,
we 
employ 
a conservative two-step heuristics \cite{sggsurvey},
which consists of
(1) 
detecting part regions (i.e., nodes),
and
(2) 
inferring 
inter-part relationships
(i.e., edges).
For the node generation step, an image of size 512$\times$512 is segmented into part regions using 
a semantic segmentation model \cite{semseg},
which consists of ResNet \cite{resnet} 
and Pyramid Pooling Module \cite{semsega} 
trained on the ADE20K dataset. 
Each semantic region is represented by 
a semantic label and a bounding box. 

To enhance the reproducibility, the semantic labels 
are re-categorized 
into 
10 
coarser 
meta-classes: ``wall," ``floor," ``ceiling," ``bed," ``door," ``table," ``sofa," ``refrigerator," ``TV," and ``Other." 
Regions with area less than 5000 pixels were considered as dummy objects and removed. 
For the edge connection step, there are two types of conditions under which a part node pair is connected by an edge. One is that the pair of bounding boxes of those parts overlap. The other is that Euclidean distance between the bounding box pair (i.e., minimum point-to-region distance) is within 20 pixels.
In preliminary experiments, we found that this heuristics often works effectivity.

To enhance the discriminativity, a spatial attribute of the part region called ``size/location" word \cite{sbow} 
is used as additional attributes of the node feature. Regarding the ``size" word, we 
classify 
the part into one of three size-words according to the area of the bounding box: 
``small (0)" $S<S_o$, 
``medium (1)" $S_o\le{S}<6S_o$, 
and 
``large (2)" $6S_o\le{S}$, 
where $S_o$ is a constant corresponding to the 1/16 of the image area. Regarding the ``location" word, we 
discretize
the center location of the bounding box by a grid of 3x3=9 cells and define the cell ID ($\in[0, 8]$) as the location word.

Note that all the attributes above are interpretable semantic words, and only complex appearance/spatial attributes 
such as real-valued descriptors 
are not introduced. Finally, the node feature is defined in the joint space of semantic, size, and location words as a 10x3x9=270 dimensional one-hot vector.

\subsection{State Recognizer}

A graph convolutional neural network GCN is employed to embed a scene graph into the state vector. The architecture of GCN is identical to that used in our recent publication \cite{icte2023ohta}. 
The implementation of GCN uses 
the deep graph library from \cite{dgl}.

In the above paper \cite{icte2023ohta}, the 2D location $(x, y)$ is used as the state representation for a passive self-localization application, ignoring the bearing attribute $\theta$. In contrast, in the active self-localization scenario considered in this work, the bearing attribute $\theta$ plays much more important role. That is, the robot should change its behavior depending on the bearing angle even when the location is unchanged. Therefore, the robot's workspace is modeled as a 3D region in the location-bearing space $(x, y, \theta)$, and it is partitioned into
a regular grid of place classes using the location and bearing resolution of 2 m and 30 deg, respectively.

Another key difference is the necessity of knowledge transfer from passive to active vision. 
The class-specific probability map output by 
GCN 
(i.e., passive vision)
is often uncalibrated as the knowledge to be transferred.
In fact, in most existing works, convolutional neural networks are used as a ranking function or classifier \cite{imagenet,actionclassifier,planet}, %
rather than a regressor \cite{regressor}. 
The class-specific rank value vector
could be naively 
used as feature to be transferred.
However, 
such a rank vector is often inappropriate as 
a feature vector because the most confident classes with highest probabilities are assigned the lowest values. 
Instead,
a reciprocal rank vector
\cite{rrf2009}
is used 
as a feature vector
and 
as the state vector
for the reinforcement learning for active vision.
Note that
reciprocal rank vector
is 
as an alternative and additive feature,
and 
proven to have
several desirable properties
in the field of multi-modal information fusion \cite{multimodal}.

\subsection{Multi- Hypothesis Tracking}

A particle filter is employed for incremental estimation of pose (location/bearing) during the sequential multi-view self-localization \cite{mcl}. 

The spatial resolution required for tracking particles in the particle filter framework typically 
needs to be 
much higher than that of the place classes \cite{dal}. 
Therefore, 
a simple max pooling is used to 
convert the particles' location attributes to the class-specific rank values, regardless of their bearing attributes. 

The particle set must be initialized at the beginning of a training/test episode. It could be naively initialized by uniformly sampling the location/bearing attributes of the particles. However, this often results in a large number of useless particles, which are lossy in both space and time. Instead, we used a guided sampling strategy \cite{guidedpf}, 
in which $M=5000$ initial particles are sampled from only those $k=3$ place classes that received the highest initial observation likelihood based on the first observation at the first viewpoint ($t=0$) in each episode. Empirically, such a guided sampling strategy 
tends to contribute 
to a significant reduction in the number of particles and computational cost with no or little loss of localization accuracy.

\subsection{Action Planner}\label{sec:planner}

Most reinforcement learning-based action planners suffer from sparse rewards as well as high-dimensionality of the state-action spaces. This issue of ``curse of dimensionality" is addressed by introducing an efficient nearest neighbor -based approximation of Q-learning (NNQL) as in \cite{nnql}.

Its training procedure is an iterative process of indexing state-action pairs associated with Q values to the database. 
Importantly,
it stores 
the Q-values 
for
only those points that are experienced in the training stage, 
rather than every possible point
in the high-dimensional state-action space,
which is intractable.
%
%
Note that the number of Q-values to be stored is significantly reduced by this strategy and becomes independent of the dimensionality.

Its action planning is a process of similarity search over the database using the queried state and averaging the $k$ Q-values that are linked to $k=4$ nearest-neighbor
state-action pairs 
$N(s, a)$ 
of the state-action pair $(s, a)$:
\begin{equation}
Q(s, a) =
|N(s, a)|^{-1}
\sum_{(s', a')\subset N(s,a)}
Q(s', a').
\end{equation}
Exceptionally, if 
there exists 
the same state-action pair 
as $(s, a)$
within the range of quantization error, 
it is regarded as revisited state-action pair and then, the Q value associated with that state-action pair is directly returned.

According to the theory of Q-learning \cite{suttonrl}, the Q function is updated by:
\begin{equation}
Q(s_t, a_t) \leftarrow Q(s_t, a_t) + \alpha
[
r_{t+1} +
\gamma \max Q(s_{t+1}, a) - Q(s_t, a_t) 
],
\end{equation}
where
$s_{t+1}$ 
is the state to which 
it is transited from the state $s_t$ by executing the action $a_t$.
$\alpha$
is the learning rate.
$\gamma$ 
is the discount factor.

It should be noted that 
this database allows the Q function 
at any past checkpoint to be recovered at low cost. 
Let $S_t$ 
and
$Q_t(s, a)$
denote 
the state-action pairs stored in the database 
and Q 
value 
linked to $(s, a)$
at $t$-th episode.
For any past checkpoint $t'(<t)$,
$S_{t'}$ is a subset of $S_t$ (i.e., $S_{t'}{\subset}S_t$).
Therefore,
Q function at checkpoint $t'$
can be recovered
from 
$Q_{t'}(s, a)$,
which 
consumes a negligible amount of storage 
compared to the 
high-dimensional state-action pairs $S_t$.
Note that 
a novel type of 
$Q_{t'}(s, a)$
``not-available (N/A)"
needs to be introduced 
for this purpose
with little storage overhead.
This allows us to store many different versions of the Q-function in memory at each checkpoint. Such memory-stored parameters are known to be beneficial in avoiding catastrophic forgetting \cite{nnforgetting}
and improved learning \cite{laine2016temporal}.

\figD

\figE

\section{Experiments}

\subsection{Dataset}

The 3D photo-realistic simulator Habitat-Sim and the dataset HM3D are used as training and deployment (test) environments, as detailed in Section \ref{sec:prob}. 
Three datasets 
called
``00800-TEEsavR23oF," ``00801-HaxA7YrQdEC," and ``00809-Qpor2mEya8F" from the Habitat-Matterport3D Research Dataset (HM3D) are considered and imported to the Habitat simulator. 
A bird's eye view of the scene and a 
first-person-view image in the environment are shown in Fig. \ref{fig:D}.

\subsection{Implementation Details}

For GCN, the number of training iterations was set as 100. The batch size was 32. The learning rate was 0.001. 

For each dataset, a GCN classifier is trained by using 10,000 training scenes with class labels as supervision.

In both training and deployment stages, the robot starts from 
a random location in the environment, performs an initial sensing, and then executes an episode consisting of a length $L=4$ sequence of plan-action-sense cycles.

The state for NNQL training and deployment at each viewpoint is represented by the latest class-specific reciprocal rank feature. The reward function returns the value of $+1$ if the Top-1 predicted place class is correct or $-1$ if it is incorrect. For NNQL training, the number of iterations is set as 10,000. The learning rate $\alpha$ of Q function is 0.1. The discount factor $\gamma$ is 0.9. The $\epsilon$-greedy with $\epsilon=(0.1(n+1)+1)^{-1}$  is used for $n$-th episode. A checkpoint was set every (1000$i$)-th episode, at which the trainable parameters were saved at each checkpoint, in a compact form as explained in Section \ref{sec:planner}. The number of test episodes was 100.

We also conducted experiments on unsupervised domain adaptation (UDA). 
The domain adaptation here evaluates the models at all checkpoints against a small validation set (size 10) from the test domain and adopts the best performing model.

The robot workspace is represented by a 3D region in the location-bearing space and it is partitioned into place classes 
by a grid-based partitioning with the location resolution of 2 m and bearing resolution of $\pi/6$ rad. 

The proposed active multi-view method 
was compared against 
the baseline 
single-view and passive multi-view methods.
The single-view method differs from the multi-view methods in its problem setting, in that no action planning is considered, but the robot performs the initial sensing and localization at 
the initial viewpoint (i.e., $t=0$) and then immediately terminates with outputting the self-localization result. The passive multi-view method is different from the 
active one
only in that the next action at each viewpoint is randomly sampled from the action space. Note that this passive multi-view framework is a standard method for passive self-localization and can be viewed as a strong baseline. In particular, as the episode length (number of observed actions per episode) increases, it is guaranteed that the performance of passive multi-view self-localization using this random action planner asymptotically approaches to the performance of best performing NBV planner.

\tabA

\subsection{Results}

For each of the three datasets, we 
evaluated
self-localization 
for 
the above three different methods 
and for 25 different settings of $T_{xy}$ and $T_{\theta}$ as described in Section \ref{sec:prob}.

Top-1 accuracy is used as the main performance index. Table \ref{tab:A} shows the performance results after 10,000 NNQL trainings. Figure \ref{fig:F} shows the results of testing for 
the experiments on unsupervised domain adaptation.

Regarding the domain adaptation results, 
early checkpoint models were also often selected as best models, in some test episodes. However, these models did not perform very well. The reason is that the model acquired in the training domain 
often already had sufficient generalization ability 
in this experiment. 
Unfortunately, under what conditions the domain adaptive model outperforms the pretrained model 
is still unclear and remains a research topic.

\figF

Figure \ref{fig:E} illustrates examples of actions and view-images in the test stage. The variable $t$ indicates the ID of plan-action-sense cycle ID, where $t=0$ corresponds to the initial sense at the start viewpoint by the robot. 

The computational cost for scene graph generation, GCN, and viewpoint planning was 18, 0.10, 10 [ms], which was faster than the real-time.


In most datasets, the passive multi-view method and the proposed method showed higher performance than the single-view method. Furthermore, in the unsupervised domain adaptation experiments, the proposed method outperforms the passive multi-view method for all the datasets considered here. Experiments showed that the proposed approach significantly improves accuracy for a wide variety of datasets.

Figure \ref{fig:E} shows success and failure examples. As can be seen from the figure, behavior of robots moving to locations with dense natural landmark objects often improves visual place recognition performance. 
For example, at the initial viewpoint, the robot was facing the wall and could not observe any effective landmarks at all, but the robot was able to detect the door by changing the heading direction at the next viewpoint, and the self-localization accuracy was improved using this landmark object. A typical failure case is also shown in Fig. \ref{fig:E}, where more than half of the viewpoints in the episode are facing feature-less objects such as walls and windows. Another notable trend is that the recognition success rate decreases when the viewpoint is too close to the object, which 
results in narrow field-of-view.

\figETM


\section{
Application: Egocentric Topological Map
}

The graph neural embedding framework introduced in this study can be considered as an instance of an egocentric topological map in recently emerged topological mapping paradigms such as graph neural SLAM \cite{pgs}.

One unresolved issue in topological navigation is active localization using topological maps. Typical frameworks for training action planners, such as reinforcement learning, assume fixed-size grid data as input (e.g., grid map), and thus they are not suitable for topological maps, which are inherently non-grid and variable-size. Moreover, their heavy computational burden will ruin the lightweight advantage of topological maps. 

To address these issues, we have been developing a novel map scheme, called "ego-centric topological map" (Fig. \ref{fig:ETM}). Unlike typical map schemes that require a globally consistent worldcentric model to be precomputed, ego-centric topological map is inherently viewpoint-specific model. It can be obtained directly from egocentric cameras, compactly compressed into graph neural network models, and transferred to action planners and planner training frameworks, 

In parallel to the work in this paper, we developed a real-time SLAM prototype for egocentric topological maps. The egocentric scene graph and graph neural network introduced in the current paper are used as the core of egocentric topological map model and mapping system, and then modules for visual recognition, memory, search, and matching are added to construct a complete prototype of topological mapping. As a result, not only visual recognition and action planning, but also planner training were achieved within the real-time budget for each viewpoint. Very recently, experiments using the photorealistic Habitat simulator validated effectiveness of the proposed approach.

\section{
Concluding Remarks
}

In this work, an entirely CPU-based lightweight framework for solving semantic localization in both active and passive 
self-localization tasks
is presented.
A scene graph and a graph neural network are combined for the first time to solve semantic localization, presenting a new graph neural localizer framework.
Experiments show that the proposed method outperforms the baseline method in terms of self-localization performance, computational efficiency, and domain adaptation.

\bibliographystyle{IEEEtran} 
\bibliography{reference}

\end{document}